\documentclass[letterpaper]{article}
\usepackage{aaai2026} 
\usepackage{times} 
\usepackage{helvet} 
\usepackage{courier} 
\usepackage[hyphens]{url} 
\usepackage{graphicx} 
\usepackage{graphicx} 
\usepackage{natbib} 
\usepackage{caption} 
\usepackage{tcolorbox}
\definecolor{myblue}{RGB}{150, 150, 230}
\tcbuselibrary{breakable} 

\usepackage{algorithm}
\usepackage{algorithmic}
\usepackage{booktabs}
\usepackage{multirow} 
\usepackage{xcolor}
\usepackage{colortbl}
\usepackage{amsmath}
\usepackage{amssymb} 
\usepackage{amsfonts} 
\usepackage{pifont}

\usepackage{newfloat}
\usepackage{listings}
\DeclareCaptionStyle{ruled}{labelfont=normalfont,labelsep=colon,strut=off} 
\floatstyle{ruled}
\newfloat{listing}{tb}{lst}{}
\floatname{listing}{Listing}

\title{
Sim4Seg: Boosting Multimodal Multi-disease Medical Diagnosis Segmentation with Region-Aware Vision-Language Similarity Masks
}
\author{
    Lingran Song,
    Yucheng Zhou,
    Jianbing Shen\thanks{Corresponding author.}
}
\affiliations{
    SKL-IOTSC, CIS, University of Macau\\
    Lingran.Song@connect.um.edu.mo, jianbingshen@um.edu.mo
}

\usepackage{bibentry}

\begin{document}

\maketitle

\begin{abstract}
Despite significant progress in pixel-level medical image analysis, existing medical image segmentation models rarely explore medical segmentation and diagnosis tasks jointly. However, it is crucial for patients that models can provide explainable diagnoses along with medical segmentation results. In this paper, we introduce a medical vision-language task named Medical Diagnosis Segmentation (MDS), which aims to understand clinical queries for medical images and generate the corresponding segmentation masks as well as diagnostic results. To facilitate this task, we first present the \textbf{M}ultimodal \textbf{M}ulti-disease \textbf{M}edical \textbf{D}iagnosis \textbf{S}egmentation (\textbf{M3DS}) dataset, containing diverse multimodal multi-disease medical images paired with their corresponding segmentation masks and diagnosis chain-of-thought, created via an automated diagnosis chain-of-thought generation pipeline. Moreover, we propose \textbf{Sim4Seg}, a novel framework that improves the performance of diagnosis segmentation by taking advantage of the \textbf{R}egion-Aware \textbf{V}ision-\textbf{L}anguage \textbf{S}imilarity to \textbf{M}ask (\textbf{RVLS2M}) module. To improve overall performance, we investigate a test-time scaling strategy for MDS tasks. Experimental results demonstrate that our method outperforms the baselines in both segmentation and diagnosis. 
\end{abstract}

\begin{links}
    \link{Code}{https://github.com/SLR567/Sim4Seg}
    \link{Dataset}{https://github.com/SLR567/M3DS}
\end{links}

\section{Introduction}

As an important clinical application, medical image segmentation aims to identify tissues, lesions, and organs in various medical images~\cite{DBLP:journals/phat/RameshKSDR21,DBLP:journals/iet-ipr/WangLCZMN22,DBLP:conf/miccai/ZhouSTL18}.
Although existing specialist models~\cite{DBLP:conf/miccai/RonnebergerFB15,DBLP:journals/mia/SahaHH21} achieve impressive performance in specific tasks, they lack the capacity to directly provide diagnosis with explanations, which is an essential capability for real-world clinical workflows.

As an extension of referring expression segmentation~\cite{DBLP:conf/eccv/HuRD16},  reasoning segmentation has recently been proposed to generate fine-grained masks for objects referenced in text output. This paradigm acquires further exploration in medical image segmentation. In the general domain, recent studies~\cite{DBLP:conf/cvpr/LaiTCLY0J24,DBLP:conf/acl/0001SS25a} have advanced fine-grained reasoning tasks for Large Vision-Language Models (LVLMs) using text prompts. These works effectively integrate visual encoders of LVLMs~\cite{DBLP:conf/nips/LiuLWL23a} with downstream task decoders~\cite{DBLP:conf/iccv/KirillovMRMRGXW23}, enhancing their visual grounding capabilities. 
Specifically, GSVA~\cite{DBLP:conf/cvpr/XiaHHPSH24} addresses the distribution gap between multiple-target and empty-target scenarios. Separately, GLaMM~\cite{DBLP:conf/cvpr/Rasheed0MS0CAX024} and PixelLM~\cite{DBLP:conf/cvpr/RenHW0FFJ24} enhance model versatility in language and vision modalities, respectively, enabling multi-granularity reasoning. READ~\cite{qian2024reasoning} leverages points as prompts to enhance segmentation performance by improving fine-grained text-image correspondence. 
Models that combine reasoning and segmentation abilities hold significant potential in clinical applications.
However, existing medical LVLMs focus primarily on segmentation capabilities~\cite{DBLP:journals/corr/abs-2504-11008} or leverage LVLMs for text-guided localization descriptions~\cite{DBLP:conf/mm/GuoCLW24,DBLP:conf/acl/0001SS25}. Developing a unified model that offers medical segmentation and explainable diagnosis capability remains an open challenge.

\begin{figure}[t]
\centering
\includegraphics[width=1\linewidth]{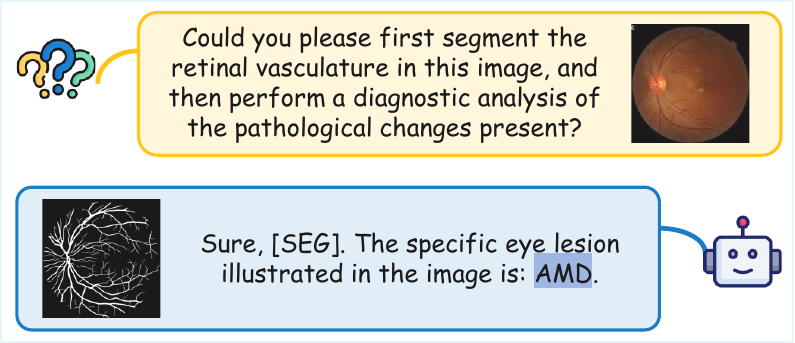}
\caption{Medical Diagnosis Segmentation (MDS) task  requires model to understand medical images and queries, then generate corresponding segmentation masks along with diagnoses.}
\label{fig:task}
\end{figure}

In this work, we propose a novel medical vision-language task, Medical Diagnosis Segmentation (MDS) (illustrated in Figure \ref{fig:task}), which requires a model to understand medical images with queries and generate both segmentation masks and corresponding diagnosis results. 
To support the MDS task and facilitate future research, we introduce the Multimodal Multi-disease Medical Diagnosis Segmentation (M3DS) dataset. M3DS contains 10 subsets across various modalities and disease types. Each sample contains an image, ground truth mask, query, diagnosis result, and a diagnosis chain-of-thought (CoT). We first select multi-modality, multi-disease medical image segmentation data from publicly available datasets. 
Subsequently, we construct CoT question-answering pairs for M3DS using our automated pipeline, which leverages the open-source medical LVLM HuatuoGPT-Vision~\cite{DBLP:conf/emnlp/ChenGOGCCWCJWW24}.
Unlike traditional medical image segmentation or medical VQA datasets~\cite{DBLP:conf/cvpr/ChengF0WLWLYCLS25,DBLP:journals/corr/abs-2305-10415}, M3DS unifies segmentation and diagnosis reasoning, enhancing pixel-level explainability for medical VQA while providing reasoning-based interpretability for segmentation, serving as a valuable resource for future research.

Building upon existing reasoning segmentation methods~\cite{DBLP:conf/cvpr/LaiTCLY0J24,DBLP:conf/iclr/LanC0XKWF025,DBLP:journals/corr/abs-2304-05653}, 
we introduce Sim4Seg. This model leverages Region-Aware Vision-Language Similarity Masks (RVLSMs) derived from text-image token embedding similarity of the last hidden state to enhance diagnosis segmentation performance.
Fine-tuning Sim4Seg on M3DS dataset improves its medical diagnosis segmentation capability. Furthermore, we develop a test-time scaling strategy especially designed for MDS task to optimize overall performance.

In summary, the main contributions of this paper are as follows:
\begin{itemize}
\item We introduce M3DS dataset, a unified resource comprising segmentation masks and diagnosis CoT generated by our automated pipeline, which facilitates future research.
\item We propose Sim4Seg, a novel framework incorporating Region-Aware Vision-Language Similarity to Mask (RVLS2M) module to enhance medical image segmentation and diagnosis capabilities.
\item We design a test-time scaling strategy specifically for medical image diagnosis and segmentation, leading to higher performance.
\item Extensive experiments demonstrate that Sim4Seg outperforms existing models on while exhibiting robust cross-dataset and cross-modality generalization capabilities.
\end{itemize}

\begin{figure*}[t]
\centering
\includegraphics[width=1\linewidth]{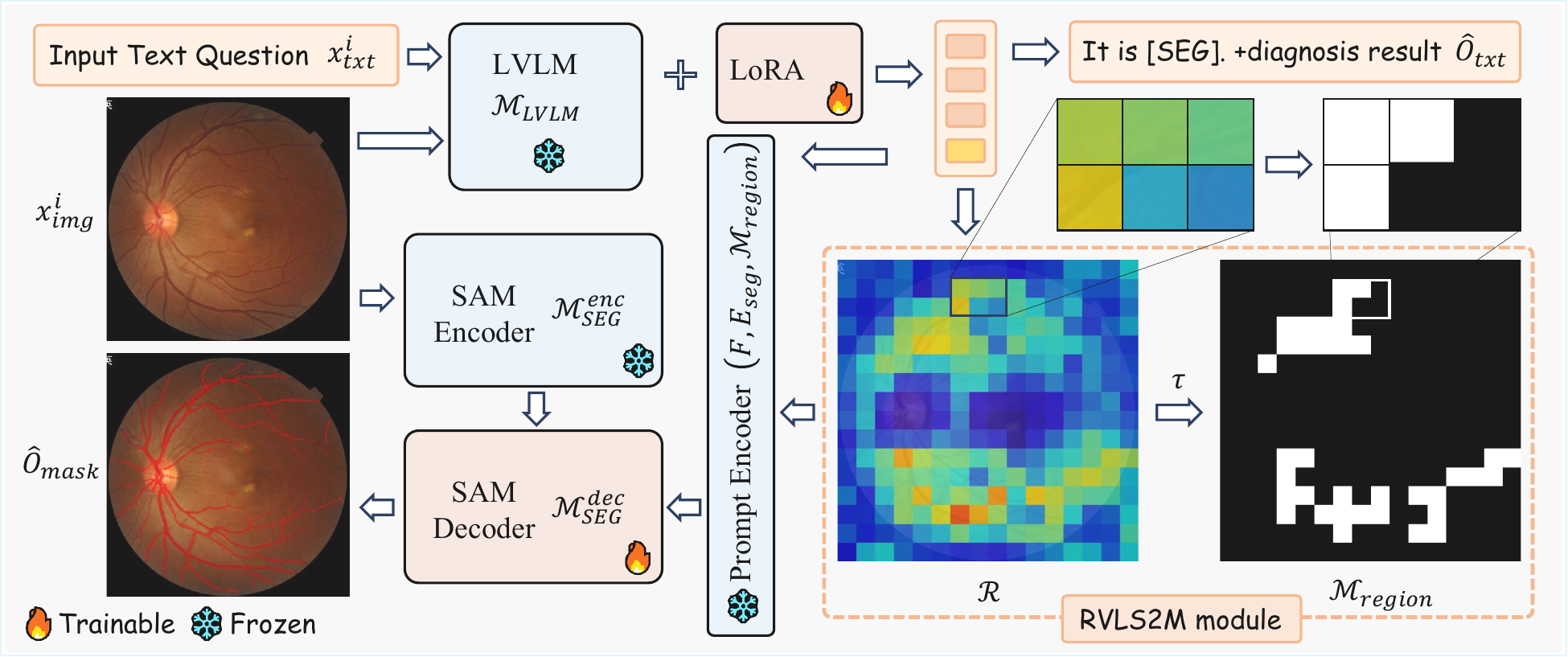}
\caption{Model Architecture of Sim4Seg. LVLM generates the corresponding output according to the given medical image and query. The output hidden states are fed into the RVLS2M module, from which $\mathbf{M}_{region}$ is obtained to prompt the SAM~\cite{DBLP:conf/iccv/KirillovMRMRGXW23} model.}
\label{fig:model_archi}
\end{figure*}

\section{M3DS Dataset}
To enhance the segmentation and diagnosis capabilities of Sim4Seg, we constructed the M3DS dataset. Firstly, we collected medical image segmentation datasets with corresponding disease categories. Subsequently, we designed a multi-role CoT data generation pipeline for M3DS dataset.

\subsection{Data Collection}
\begin{table}[t]\small
\centering
\caption{Overview of \textbf{M3DS} dataset. In particular, \textbf{DS} stands for Dermoscopy, \textbf{End} refers to Endoscopy, \textbf{US} denotes Ultrasound, \textbf{FP} represents Fundus Photography, and \textbf{BFD} signifies Bone Fracture Detection.}
\label{tab:dataset}
\setlength{\tabcolsep}{3pt}
\begin{tabular}{l|ccccc|ccc}
\toprule
\bf \multirow{2}{*}{Dataset}        & \multicolumn{5}{c|}{\textbf{Modality}}& \multicolumn{3}{c}{\textbf{Size}}\\
\cmidrule(lr){2-6}\cmidrule(lr){7-9}&\bf X-Ray&\bf DS&\bf End&\bf US&\bf FP& \bf Train & \bf Val & \bf Test \\
\midrule
FracAtlas &\cellcolor{gray!15} \ding{51}& & & & & 574 & 82 & 61 \\
bone fracture & \cellcolor{gray!15}\ding{51}& & & & & 311 & 83 & 43 \\
BFD & \cellcolor{gray!15}\ding{51}& & & & & 1804 & 173 & 83 \\
ISBI & &\cellcolor{gray!15}\ding{51} & & & & 899 & 277 & 101 \\
ISIC & &\cellcolor{gray!15}\ding{51} & & & & 2000 & 150 & 600 \\
Kvasir-SEG & & &\cellcolor{gray!15}\ding{51} & & & 801 & 99 & 100 \\
BUSI & & & &\cellcolor{gray!15}\ding{51} & & 532 & 55 & 60 \\
TN3K & & & &\cellcolor{gray!15}\ding{51} & & 2001 & 878 & 614 \\
ChestX-Det & \cellcolor{gray!15}\ding{51}& & & & & 2478 & 388 & 101 \\
FIVES & & & & &\cellcolor{gray!15}\ding{51} & 600 & 99 & 101 \\\midrule
\textbf{Total} &4&2 &1 &2 &1 & 12000 & 2284 & 1864 \\
\bottomrule
\end{tabular}
\end{table}

We constructed M3DS by integrating ten distinct sub-datasets from diverse sources.
Specifically, FracAtlas~\cite{Abedeen_2023} contains 4,083 X-Ray images, including 717 fracture cases, each annotated with polygonal segmentation masks.
Bone fracture~\cite{bone-fracture-7fylg_dataset} comprises 458 X-Ray images, with 437 containing fracture regions.
Bone Fracture Detection (BFD)~\cite{bfd} includes 2,060 fracture X-Ray images with pixel-level segmentation masks at six anatomical sites.
ISBI~\cite{DBLP:journals/corr/GutmanCCHMMH16} and ISIC~\cite{DBLP:conf/isbi/CodellaGCHMDKLM18} contain 1,279 and 2750 images of benign or malignant skin lesions with corresponding segmentation masks, respectively.
Kvasir-SEG~\cite{DBLP:conf/mmm/JhaSRHLJJ20} provides 1,000 polyp images paired with segmentation masks.
BUSI~\cite{ALDHABYANI2020104863} comprises 780 breast ultrasound images with 437 benign, 210 malignant and 133 normal cases. 
TN3K~\cite{DBLP:journals/cbm/GongCCLLC23} includes 3,493 thyroid nodule images with corresponding  masks.
ChestX-Det~\cite{lian2021structure} contains 3,578 images sourced from NIH ChestX-14~\cite{DBLP:conf/cvpr/WangPLLBS17}, annotated by three radiologists across 13 abnormality categories.
FIVES~\cite{FIVES} features 800 fundus photographs with manually annotated masks.
The modality of these data varied from X-Ray, dermoscopy, endoscopy, ultrasound, to fundus photography. Specific details regarding the segments used from each sub-dataset are summarized in Table \ref{tab:dataset}.

\subsection{Chain-of-Thought Reasoning Data Generation}\label{sec:cot_gen}
\begin{figure*}[t]
\centering
\includegraphics[width=1\linewidth]{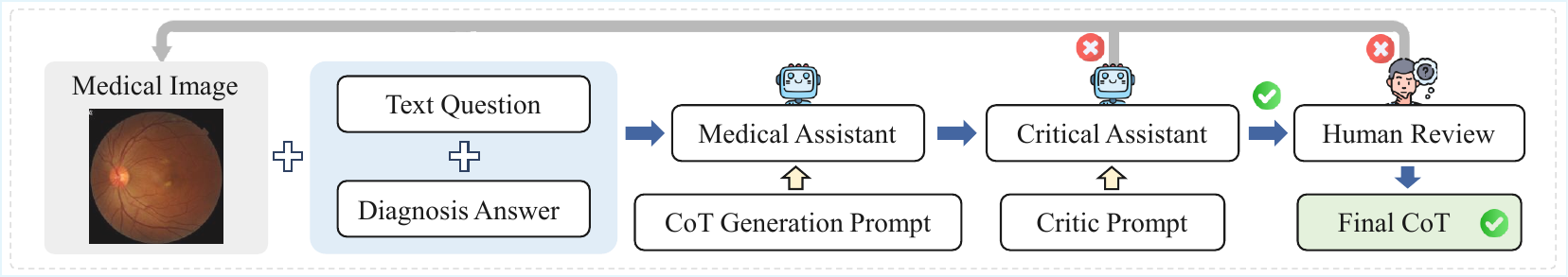}
\caption{CoT generation pipeline for M3DS dataset construction.}
\label{fig:cot}
\end{figure*}

To construct the M3DS dataset, we designed a multi-role CoT diagnosis data generation pipeline specifically for MDS tasks, leveraging HuatuoGPT-Vision~\cite{DBLP:conf/emnlp/ChenGOGCCWCJWW24} model as medical assistant and critical assistant. As illustrated in Figure \ref{fig:cot}, we first assemble the collected medical images, the corresponding questions, and the diagnosis results into a structured prompt. To guide the model toward a step-by-step understanding of the image and diagnosis reasoning, our prompt instructs the medical assistant to begin by identifying the image modality, then progressively analyze the medical image and finally derive the diagnosis. This prompt is then fed into the medical assistant to generate the CoT of diagnosis reasoning.
To ensure the quality and reliability of the generated result, a review step is incorporated into the pipeline, where the critical assistant evaluates the output. 
If the generated CoT is rejected by the critical assistant within the maximum allowed review rounds, the failure feedback is sent back to the first step to trigger a new output.
To further guarantee the effectiveness of the generation, we added a human-assisted review phase after critical assistant evaluation. 
Using this multi-role diagnosis CoT generation pipeline, we train the Sim4Seg model on our constructed M3DS dataset.

\section{Methodology}
In this section, we present our model architecture in Section \ref{sim4segarchi}, which contains our proposed Region-Aware Vision-Language Similarity to Mask (RVLS2M) module. Then, we introduce training objectives in Section \ref{trainingobj}, and a test-time scaling strategy designed for MDS task in Section \ref{tts4mds}.
\subsection{Sim4Seg Model Architecture}\label{sim4segarchi}
Let $X_{img} \in \mathbb{R}^{n \times h \times w \times c}$ be the set of input images, with $\mathrm{x}^{i}_{img} \in X_{img}$ representing a single image. Here $n$, $h$, $w$, and $c$ denote the image count, height, width, and channel count, respectively. The corresponding set of text inputs, $X_{txt}$ representing text queries, contains elements $\mathrm{x}^{i}_{txt} \in X_{txt}$. The primary task of MDS task is to predict text output $\hat{O}_{txt}$ and its corresponding segmentation mask $\hat{O}_{mask}$ as
\begin{equation*}
\Theta _{MLE}  =\mathop{\arg\max}\limits_{\Theta}\mathcal{M} _{\theta }\left (\hat{O}_{txt},\hat{O} _{mask}\mid X_{img},X_{txt};\Theta  \right ),\label{def_seg}
\end{equation*}
where the medical diagnosis segmentation model is denoted by $\mathcal{M}_{\theta}$, and $\Theta$ represents its parameters. As shown in Figure \ref{fig:model_archi}, this framework comprises an LVLM $\mathcal{M}_{LVLM}$ and a segmentation backbone $\mathcal{M}_{SEG}$, formally defined as $\mathcal{M}_{\theta} = \mathcal{M}_{LVLM} \oplus \mathcal{M}_{SEG}$. Here, $\oplus$ indicates a cascading operation between modules.
The output segmentation mask $\hat{O}_{mask} \in \{0,1\}^{h\times w}$ is a binary matrix.
To enable $\mathcal{M}_{\theta}$ to generate mask embeddings, LISA~\cite{DBLP:conf/cvpr/LaiTCLY0J24} expands the text vocabulary of $\mathcal{M}_{LVLM}$ by introducing a special token. The input image $\mathrm{x}^{i}_{img}$ is divided into uniform patches and processed by CLIP~\cite{DBLP:conf/icml/RadfordKHRGASAM21} encoder.
During training, this special token is incorporated into $\mathrm{x}^{i}_{txt}$, and both $\mathrm{x}^{i}_{img}$ and $\mathrm{x}^{i}_{txt}$ are put into $\mathcal{M}_{LVLM}$, producing language response $\hat{O}_{txt}$ as
\begin{align}
\hat{O}_{txt}=\mathcal{M}_{LVLM}\left(\mathrm{x}^{i}_{img},\mathrm{x}^{i}_{txt}\right).\label{def_txt}
\end{align}
Before predicting the binary segmentation mask, $\mathcal{M}_{\theta}$ generates a response $\hat{O}_{txt}$ containing the special token representing the target object. Following LISA~\cite{DBLP:conf/cvpr/LaiTCLY0J24}, the last hidden layer embedding $\tilde{\mathbf{E}}_{seg}$ associated with this special token is extracted from $\mathcal{M}_{LVLM}$. This is subsequently projected through a multilayer projection layer $\phi$ to obtain the refined feature $\mathbf{E}_{seg}$ as
\begin{align}
\mathbf{E}_{seg}=\phi\left(\tilde{\mathbf{E}}_{seg}\right).\label{def_Eseg}
\end{align}
Meanwhile, the visual backbone $\mathcal{M}^{enc}_{SEG}$ extracts visual features $\mathbf{F}$ from the input image $\mathrm{x}^{i}_{img}$, formulated as
\begin{align}
\mathbf{F}=\mathcal{M}^{enc}_{SEG}\left(\mathrm{x}^{i}_{img}\right).\label{def_F}
\end{align}
Building on the capability of SAM~\cite{DBLP:conf/iccv/KirillovMRMRGXW23}, which supports various types of prompts, we investigate region-aware masks as prompts via region-aware vision-language similarity. Formally, let $\mathbf{E} =\{\mathbf{E}_{1},\dots,\mathbf{E}_{n}\mid \mathbf{E}_{i} \in \mathbb{R}^d \}$ denote the hidden states of $\mathcal{M}_{LVLM}$, where $n$ is the token count, and $d$ denotes the embedding dimension. $\mathbf{E}_{seg}$ in Equation \ref{def_Eseg} satisfies $\mathbf{E}_{seg} \in \mathbf{E}$. During training, visual features and text instruction $\mathrm{x}^{i}_{txt}$ are jointly used as input to LVLM. Consequently, it has $\mathbf{E}_{img} \subseteq \mathbf{E}$. The vision-language similarity score is defined as
\begin{align}
\mathrm{Sim}=\mathbf{E}_{img}\cdot \left(\mathbf{E}_{seg}\right)^{\mathrm{T}},\label{def_simScore}
\end{align}
where $\mathrm{Sim} \in \mathbb{R}^{n}$ measures the similarity between each image token and the special token generated by LVLM.
Along with $\mathbf{E}_{seg}$ in Equation \ref{def_Eseg} and $\mathrm{Sim}$ defined in Equation \ref{def_simScore}, we propose the Region-Aware Vision-Language Similarity to Mask (RVLS2M) module detailed in Algorithm \ref{alg:RVLS2M} for improving segmentation capabilities.
After calculating the similarity score $\mathrm{Sim}$, we normalize the similarity scores via softmax to enhance the separability of regions as follows
\begin{align}
\mathrm{Sim}_{\mathrm{norm}} = \mathrm{softmax}(\mathrm{Sim}) = \left\{ \frac{\exp(s_i)}{\sum_{j=1}^{n}\exp(s_j)} \right\}_{i=1}^{n}, \label{eq:norm}
\end{align}
where $s_i$ denotes the $i$-th element in $\mathrm{Sim}$.
Next, we construct a region-aware vision-language similarity map. The normalized scores $\mathrm{Sim}_{\mathrm{norm}}$ are reshaped into a 2D map $M \in \mathbb{R}^{h' \times w'}$, with dimensions $h' = \lfloor \sqrt{n} \rfloor$ and $w' = \lceil n / h' \rceil$, such that
\begin{align}
M_{u,v} = \mathrm{Sim}_{\mathrm{norm}}[u \cdot w' + v],
\end{align}
where $\forall u \in [0, h'-1], v \in [0, w'-1]$.
This map is then divided into non-overlapping $g \times g$  grids. Within each grid cell, we compute region similarity $\mathcal{R}_{k,l}$ by average pooling
\begin{align}\label{eq:region_sim}
\begin{split}
&\mathcal{R}_{k,l} = \frac{1}{b^2} \sum_{i=bk}^{b(k+1)-1} \sum_{j=bl}^{b(l+1)-1} M_{i,j} ,
\end{split}
\end{align}
with parameters $b = \lfloor \min(h', w') / g \rfloor$, $k,l \in [0, g-1]$. $\mathcal{R} \in \mathbb{R}^{g \times g}$ is the region-aware vision-language similarity matrix.
Then, a binary segmentation mask $\mathbf{M}_{region}$ is generated by applying adaptive thresholding to $\mathcal{R}$, such that
\begin{align}
\mathbf{M}_{region} = \mathbb{I}\left( \tau(\mathcal{R}) \right), \label{eq:mask_strategy}
\end{align}
where $\mathbb{I}(\cdot)$ is the indicator function. The specific selection of the $\tau$-strategy will be detailed in Section \ref{sec:exp}. Finally, the binary segmentation mask $\mathbf{M}_{region}$ along with the special  token embedding and visual features $\mathbf{F}$ from the input image $\mathrm{x}^{i}_{img}$ are fed into $\mathcal{M}^{dec}_{SEG}$, formulated as
\begin{align}
\hat{O} _{mask}=\mathcal{M}^{dec}_{SEG} (\mathbf{F},\mathbf{E}_{seg},\mathbf{M}_{region}).\label{eq:final_output}
\end{align}

\begin{algorithm}[t]
\caption{Region-Aware Vision-Language Similarity to Mask (RVLS2M)}
\label{alg:RVLS2M}
\begin{algorithmic}[1]
\REQUIRE 
Image tokens $\mathbf{E}_{img} \in \mathbb{R}^{n \times d}$, special token embedding $\tilde{\mathbf{E}}_{seg} \in \mathbb{R}^{d}$, projection function $\phi$, grid size $g.$
\ENSURE 
Binary mask $\mathbf{M}_{region} \in \{0,1\}^{g \times g}.$
\STATE $\mathbf{E}_{seg} \gets \phi(\tilde{\mathbf{E}}_{seg})$ 
\STATE $\mathrm{Sim} \gets \mathbf{E}_{img} \cdot (\mathbf{E}_{seg})^{\mathrm{T}}$
\STATE $\mathrm{Sim}_{\mathrm{norm}} \gets \mathrm{softmax}(\mathrm{Sim})$
\STATE $h' \gets \lfloor \sqrt{n} \rfloor,\ w' \gets \lceil n / h' \rceil$
\STATE $M \gets \mathrm{reshape}(\mathrm{Sim}_{\mathrm{norm}}, [h', w'])$
\STATE $b \gets \lfloor \min(h', w') / g \rfloor$
\FOR{$k \gets 0$ \TO $g-1$}
    \FOR{$l \gets 0$ \TO $g-1$}
        \STATE $\mathcal{R}_{k,l} \gets \mathrm{mean}(M[bk:b(k+1), bl:b(l+1)])$
    \ENDFOR
\ENDFOR
\STATE $\mathbf{M}_{region} \gets  \mathbb{I}\left( \tau(\mathcal{R}) \right)$
\RETURN $\mathbf{M}_{region}$
\end{algorithmic}
\end{algorithm}

\subsection{Training Objectives}\label{trainingobj}
By jointly optimizing the text generation loss $\mathcal{L} _{txt}$ in $\mathcal{M}_{LVLM}$ and the segmentation mask loss $\mathcal{L} _{mask}$ in $\mathcal{M}_{SEG}$, we enable $\mathcal{M}_{\theta}$ to perform MDS task.
For $\mathcal{L} _{txt}$, we employ cross-entropy loss, and for $\mathcal{L} _{mask}$, we use a combination of binary cross-entropy (BCE) loss $\mathcal{L}_{bce}$ and DICE loss $\mathcal{L}_{dice}$. Finally, the overall loss objective $\mathcal{L}$ is the weighted average of $\mathcal{L}_{txt}$ and $\mathcal{L}_{mask}$, denoted by
\begin{align}\label{eq:loss}
\begin{split}
&\mathcal{L}_{mask}=\lambda _{bce}\mathcal{L}_{bce}+\lambda _{dice}\mathcal{L}_{dice},\\
&\mathcal{L}=\lambda _{txt}\mathcal{L}_{txt}+\lambda _{mask}\mathcal{L}_{mask},
\end{split}
\end{align}
where $\lambda _{bce}$ and $\lambda _{dice}$ are the weight of BCE loss $\mathcal{L}_{bce}$ and DICE loss $\mathcal{L}_{dice}$, respectively. The overall loss objective $\mathcal{L}$ is weighted by $\lambda _{txt}$ for $\mathcal{L}_{txt}$ and $\lambda _{mask}$ for $\mathcal{L}_{mask}$.

\begin{figure}[t]
\centering
\includegraphics[width=1\linewidth]{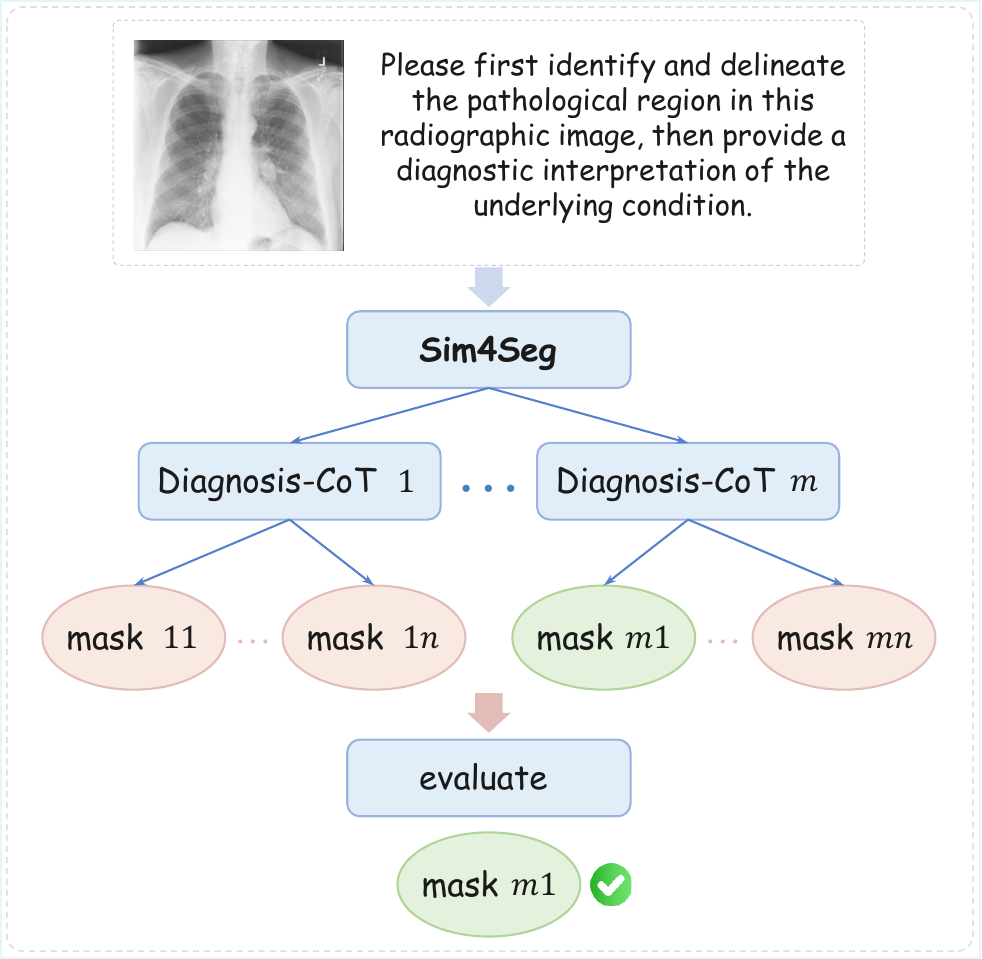}
\caption{Test-Time Scaling strategy for MDS task.}
\label{fig:TTS}
\end{figure}

\subsection{Test-Time Scaling for Medical Diagnosis Segmentation Task}\label{tts4mds}

To enhance model performance during inference, we propose a test-time scaling strategy designed for MDS task that leverages multi-path reasoning from LVLMs. Given an input medical image $\mathrm{x}_{img}$ and its corresponding query $\mathrm{x}_{txt}$, our method generates diverse outputs through a two-stage reasoning process.
Firstly, LVLM $\mathcal{M}_{LVLM}$ generates $m$ diverse diagnosis reasoning paths, represented by
\begin{align}\label{equ:tts_m}
\{O^{i}_{txt}\}_{i=1}^m = \mathcal{M}_{LVLM}(\mathrm{x}_{img}, \mathrm{x}_{txt}),
\end{align}
where each $O^{i}_{txt}\in \mathbb{R}^d$ represents one of the $m$ CoT reasoning output paths.
For each path $i$, we extract the last hidden layer embedding $\tilde{\mathbf{E}}_{seg}^i$ corresponding to the special token in $O^{i}_{txt}$, and project it to obtain the refined feature $\mathbf{E}_{seg}^i$. Combined with image tokens $\mathbf{E}_{img}$ derived from the vision encoder, binary region masks $\mathbf{M}^{i}_{region}$ are generated through the RVLS2M module.
Then, the segmentation model $\mathcal{M}_{SEG}$ produces $m\times n$ masks, formulated as
\begin{align}\label{equ:tts_n}
O^{i,j}_{mask} = \mathcal{M}_{SEG}\left(\mathrm{x}_{img}, \mathbf{E}_{seg}^i,g(\mathbf{M}^{i}_{region}, \theta_j)\right),
\end{align}
with $j=1,\dots,n$, and $g(\cdot)$ denoting a stochastic perturbation parameterized by $\theta_j \sim \mathcal{T}$ to ensure diversity in mask generation.
The final segmentation mask is selected by maximizing an evaluation metric $\mathcal{Q}$ as follows
\begin{align}\label{equ:tts_all}
O_{mask}^{final} = \underset{O^{i,j}_{mask}}{\arg\max}  \mathcal{Q}\left(O^{i,j}_{mask}, \hat{O}_{mask}\right),
\end{align}
where $\mathcal{Q}$ is the quality metric computed as the average of gIoU and cIoU.
As illustrated in Figure \ref{fig:TTS}, this strategy generates $m \times n$ candidate masks and selects one based on evaluation performance.

\section{Experiments}\label{sec:exp}

\begin{table}[t]
\centering
\caption{Main results on test set of \textbf{M3DS} dataset.``ft'' refers to fine-tuning model with non-diagnostic settings, ``ft-diagnosis'' indicates fine-tuning model with diagnosis option but without chain-of-thought data, and ``ft-CoT'' denotes fine-tuning model with chain-of-thought data.}
\label{tab:main}
\setlength{\tabcolsep}{3pt}
\begin{tabular}{lccc}
\toprule
\bf \multirow{2}{*}{Method} &  \multicolumn{3}{c}{\textbf{overall}} \\ \cmidrule{2-4}       & \bf gIoU & \bf cIoU & \bf Acc  \\
\midrule
LLaVA-Med~\cite{DBLP:conf/nips/LiWZULYNPG23} & -  & - & 3.48    \\
SAM-Med2D~\cite{DBLP:journals/corr/abs-2308-16184} & 22.94  & 51.42 & -   \\
READ~\cite{qian2024reasoning} & 13.37 & 25.75 & 2.52 \\
LISA~\cite{DBLP:conf/cvpr/LaiTCLY0J24} & 32.43  & 31.83 & 4.71    \\
LISA (ft)~\cite{DBLP:conf/cvpr/LaiTCLY0J24}    & 44.07   & 42.89   & 0.00 \\
LISA (ft-diagnosis)~\cite{DBLP:conf/cvpr/LaiTCLY0J24}    & 45.87   & 46.05   & 53.27 \\
LISA (ft-CoT)~\cite{DBLP:conf/cvpr/LaiTCLY0J24}    & 45.90   & 45.92   & 58.05 \\
Sim4Seg (ft-diagnosis)   & 51.00   & 54.06   & 54.33 \\
Sim4Seg (ft-CoT)   & 51.86   & 53.90   & 69.04 \\
\rowcolor{gray!15}Sim4Seg (ft-CoT) +test-time scaling       & \bf53.11  & \bf55.83         & \bf82.63   \\
\bottomrule
\end{tabular}
\end{table}

\subsection{Experimental Setting}
\subsubsection{Dataset and Metric.}
In our experiments, we trained Sim4Seg model on the training split (12,000 samples) of the proposed M3DS dataset and evaluated it on the test split (1,864 samples). Accuracy (Acc) was employed to evaluate the accuracy of diagnosis results. 
gIoU~\cite{DBLP:conf/cvpr/RezatofighiTGS019} and cIoU~\cite{DBLP:conf/aaai/ZhengWLLYR20} measure the overlap between the predicted segmentation masks and their corresponding ground truth. 

\subsubsection{Implementation Details.}
In the experiments, we employ LISA~\cite{DBLP:conf/cvpr/LaiTCLY0J24} to initialize our model. During training, we trained for four epochs, employed an AdamW optimizer~\cite{DBLP:conf/iclr/LoshchilovH19} with a learning rate of $3 \times 10^{-4}$, a weight decay of $0.01$, a batch size of $2$, and a gradient accumulation step of $10$. The default loss weights $\lambda_{mask}$ and $\lambda_{txt}$ are set to $1.0$, the BCE loss weight $\lambda_{bce}$ is set to $2.0$ and the DICE loss weight $\lambda_{dice}$ is set to $0.5$. All experiments were conducted on an NVIDIA H800 GPU.

\begin{figure}[t]
\centering
\includegraphics[width=1\linewidth]{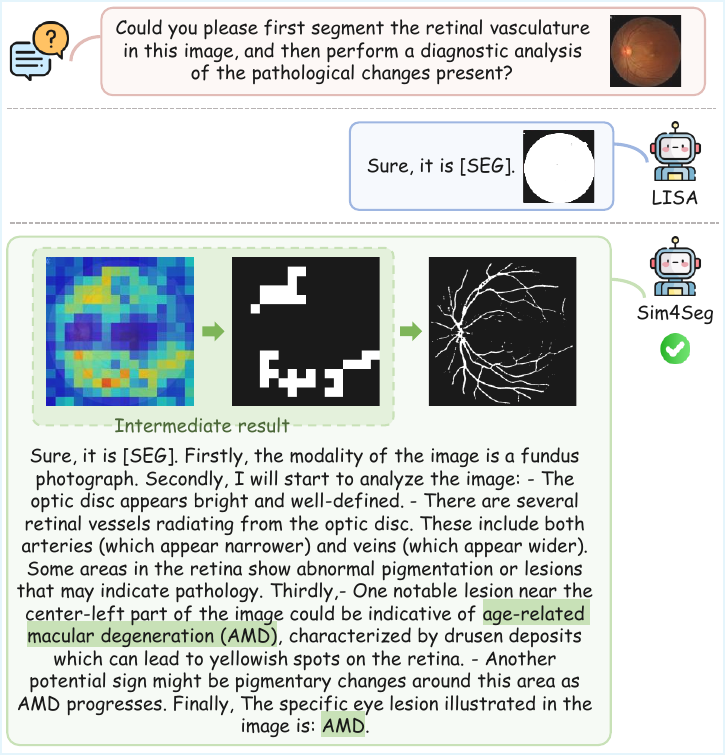}
\caption{Case Study.}
\label{fig:case}
\end{figure}

\subsection{Main Results on M3DS Dataset}
The proposed Sim4Seg model demonstrates state-of-the-art performance on the M3DS dataset (5 modalities, 10 sub-datasets) through comprehensive experiments. 
As shown in Table \ref{tab:main}, Sim4Seg exhibits significant improvements on evaluation metrics. 
Sim4Seg enhances medical vision-language models with segmentation capabilities while integrating medical diagnosis into segmentation models.
Sim4Seg exceeds reasoning segmentation models by +57.3\% segmentation performance and +165.4\% diagnosis accuracy.
\subsection{Ablation Study}
\begin{table*}[t]\small
\centering
\caption{Detailed ablation result on each sub-set of \textbf{M3DS} dataset. \textbf{BF} refers to bone fracture dataset, \textbf{BFD} indicates Bone Fracture Detection dataset, \textbf{KS} denotes Kvasir-SEG dataset, and \textbf{CXD} represents ChestX-Det dataset.}
\label{tab:ab}
\setlength{\tabcolsep}{5pt}
\begin{tabular}{l|c|cccccccccc|c}
\toprule
 \multicolumn{2}{l|}{\textbf{Method}}     & \bf FracAtlas & \bf BF  & \bf BFD & \bf ISBI & \bf ISIC & \bf KS & \bf BUSI & \bf TN3K & \bf CXD & \bf FIVES & \bf Avg. \\
\midrule
 \rowcolor{gray!20} \multicolumn{13}{c}{\textit{w/o  RVLS2M}}      \\
\midrule
\multirow{3}{*}{zero-shot} & gIoU & 1.42  & 3.42 & 6.26  & 81.25  & 47.71  & 45.49  & 41.70  & 23.56  & 5.93  & 7.47  & 26.42  \\
&cIoU & 1.43 &2.92 &5.99 &76.69 &49.21 &46.74 &32.20 &21.67 &6.99 &7.21 &25.11 \\
&Acc & 28.36 &10.45 &0.00 &9.9 &0.00 &32.00 &36.67 &0.00 &0.00 &1.98 &11.94 \\
\midrule
 \multirow{3}{*}{+FT w/o diagnosis} & gIoU & 1.63 &5.15 &8.35 &85.11 &63.40 &47.03 &42.58 &37.79 &11.99 &27.90 &33.09  \\
&cIoU & 1.50 &3.31 &10.00 &82.26 &62.71 &46.08 &45.21 &33.90 &18.84 &28.90 &33.27 \\
&Acc & 0.00  & 0.00 & 0.00  & 0.00  & 0.00  & 0.00  & 0.00  & 0.00  & 0.00  & 0.00  & 0.00 \\
\midrule
 \multirow{3}{*}{+FT w/ diagnosis} & gIoU & 1.50
&5.23 &10.28 &85.93 &63.59 &56.44 &41.91 &40.01 &14.69 &32.55
&\textbf{35.21} \\
&cIoU & 1.69 &7.10 &11.00 &83.50 &63.25 &54.48 &46.46 &39.62  &21.34 &33.86 &\textbf{36.23} \\
&Acc & 88.06 &79.10 &48.19 &69.31 &64.50 &82.00 &63.33 &34.20 &36.63 &32.67 &59.80 \\
\midrule
 \multirow{3}{*}{+FT w/ diagnosis-CoT} & gIoU & 1.58
&5.15 &8.35 &85.09 &63.97 &54.29 &45.91 &40.12 &15.26 &31.89 &35.16 \\
&cIoU & 1.71 &3.31 &10.00 &81.85 &64.85 &49.79 &47.30 &38.93 &22.49 &33.22 &35.35 \\
&Acc & 88.52 &81.39 &57.83 &73.27 &64.50 &84.00 &80.00 &45.60 &39.60 &31.68 &\textbf{64.64} \\
\midrule
 \rowcolor{gray!20} \multicolumn{13}{c}{\textit{w/  RVLS2M}}      \\
\midrule
\multirow{3}{*}{+FT w/ diagnosis} & gIoU & 3.39
&6.99 &10.85 &86.26 &69.49 &65.08 &50.72 &46.38 &19.17 &32.44 &39.08  \\
&cIoU & 3.99 &7.72 &11.51 &83.74 &71.88 &59.17 &52.23 &51.36 &35.84 &34.17 &41.16 \\
&Acc & 88.06 &82.09 &42.17 &54.46 &63.83 &57.00 &58.33 &46.09 &34.65 &31.68 &55.84 \\
\midrule
 \multirow{3}{*}{+FT w/ diagnosis-CoT} & gIoU & 6.41
&7.07 &11.97 &86.96 &68.37 &67.23 &49.42 &48.31 &20.98 &36.70 &40.34  \\
&cIoU & 5.42 &5.46 &13.86 &85.47 &67.02 &60.17 &49.49 &54.18 &37.13 &38.64 &41.68 \\
&Acc & 89.55 &86.57 &63.86 &74.26 &68.00 &87.00 &81.67 &72.80 &41.58 &33.66 &69.90 \\
\midrule
  \multirow{3}{*}{+FT w/ diagnosis-CoT (TTS)} & gIoU & 6.98 &7.48 &13.61 &87.44 &69.33 &68.28 &49.28 &50.03 &23.53 &37.77 &\textbf{41.37}  \\
 &cIoU & 6.21 &9.40 &15.90 &85.48 &69.72 &63.04 &50.10 &55.66 &39.58 &39.75 &\textbf{43.48} \\
 &Acc & 98.51 &95.52 &80.72 &80.20 &76.83 &92.00 &78.33 &93.81 &45.54 &68.32 &\textbf{80.98} \\
\bottomrule
\end{tabular}
\end{table*}

We evaluated Sim4Seg components by testing individual modules and measuring their performance effects. Table \ref{tab:ab} compares results w/ and w/o RVLS2M.
Overall, the w/ RVLS2M configuration consistently outperforms the w/o RVLS2M setup. FT w/o diagnosis (fine-tuning without diagnosis text like ``It is [SEG]") improves medical image segmentation capability but lacks diagnosis capabilities. FT w/ diagnosis (fine-tuning with diagnosis text formatted as ``It is [SEG]. + diagnosis result") enhances both segmentation and diagnosis performance.
FT w/ diagnosis-CoT (fine-tuning with diagnostic chain-of-thought) adopts format ``It is [SEG]. + diagnosis CoT", with CoT generated as described in Section \ref{sec:cot_gen}, significantly improves diagnosis performance. In summary, training on M3DS dataset enhances segmentation and diagnosis performance. 
Our proposed test-time scaling (TTS) strategy also improves both segmentation and diagnosis performance, indicating that every component contributes critically to the MDS task.

\subsection{Analysis}

\subsubsection{Impact of RVLS2M Module Under Zero-Shot Setting.}

\begin{table}[t]
\centering
\caption{Zero-shot effectiveness of RVLS2M module. The proposed RVLS2M module also serves as an efficient prompt creator in zero-shot scenarios.}
\label{tab:RVLS2M}
\setlength{\tabcolsep}{5pt}
\begin{tabular}{lccc}
\toprule
\bf Method & \bf gIoU & \bf cIoU & \bf Avg. \\
\midrule
LISA~\cite{DBLP:conf/cvpr/LaiTCLY0J24} & \bf 32.43 & 31.83 & 32.13 \\
 \rowcolor{gray!15}LISA w/ RVLS2M module & 31.82 & \bf 39.88 & \bf 35.85\\
\bottomrule
\end{tabular}
\end{table}

The proposed RVLS2M module demonstrates significant  improvements in segmentation performance, as validated in ablation studies.
Notably, it also serves as an effective prompt creator in zero-shot scenarios. As shown in Table \ref{tab:RVLS2M}, using LISA~\cite{DBLP:conf/cvpr/LaiTCLY0J24} as the base model and incorporating the RVLS2M module without any training improves performance by 11.6\%. This confirms the plug-and-play effectiveness and flexibility of the RVLS2M module.

\subsubsection{Impact of Different $\tau$ Strategies.}

\begin{figure}[t]
\centering
\includegraphics[width=1\linewidth]{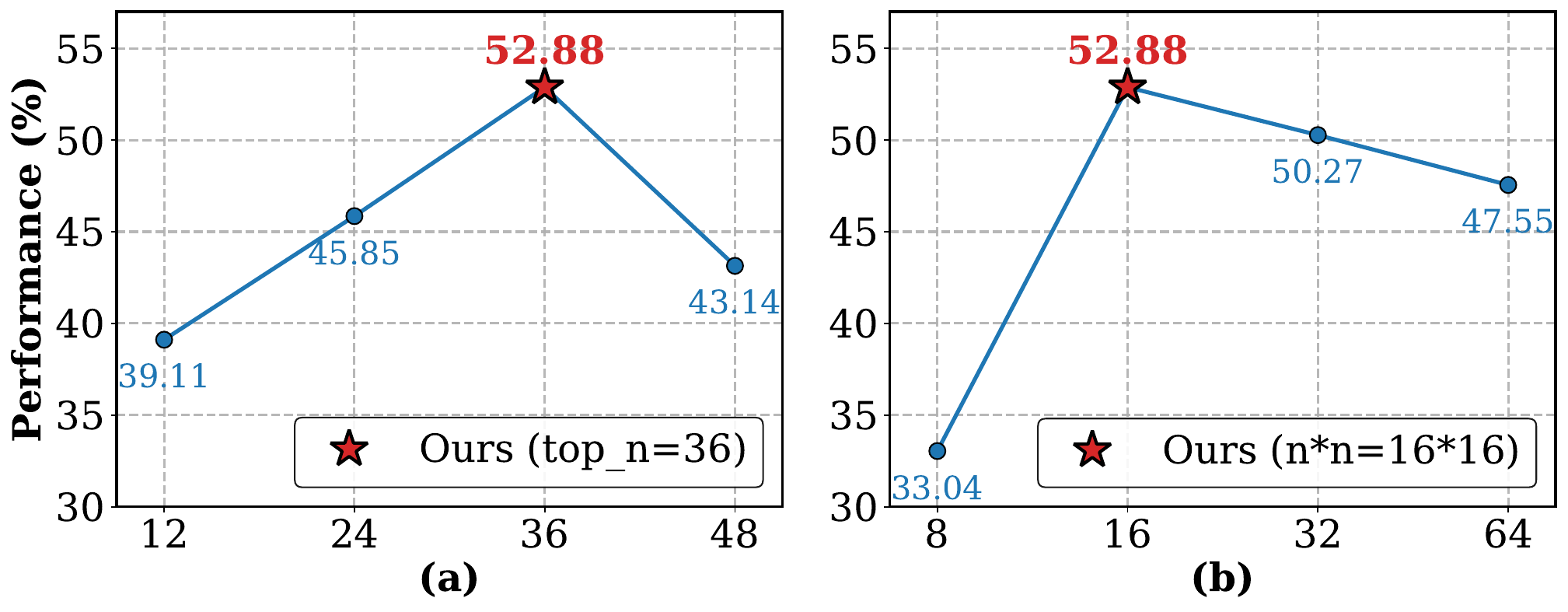}
\caption{Impact of Different $\tau$ Strategies. An inverted U-shaped relationship exists between segmentation performance and RVLSM granularity, controlled by $\tau$. Selecting 36 grid cells (a) and using a $16\times16$ grid resolution (b) achieve peak performance.}
\label{fig:top_n}
\end{figure}

Figure \ref{fig:top_n} reveals an inverted U-shaped relationship between segmentation performance and RVLSM granularity, controlled by the $\tau$ strategy illustrated in Equation \ref{eq:mask_strategy}. Performance improves when selecting 12 to 36 grid cells with highest similarity, but declines at 48 grid cells. 
Similarly, optimal results emerge when selecting the top 36 grid cells at $16\times 16$ grid resolution, with performance increasing from $8\times 8$ to $16\times 16$ grid resolution and decreasing from $16\times 16$ to $64\times 64$ grid resolution.
This indicates that excessively coarse grids produce blurred regions, while overly fine grids retain spurious correlations from high-similarity points. These findings demonstrate the critical role of optimal $\tau$ strategy selection in RVLSM effectiveness.

\subsubsection{Impact of Different Test-Time Scaling Strategies.}

\begin{figure}[t]
\centering
\includegraphics[width=1\linewidth]{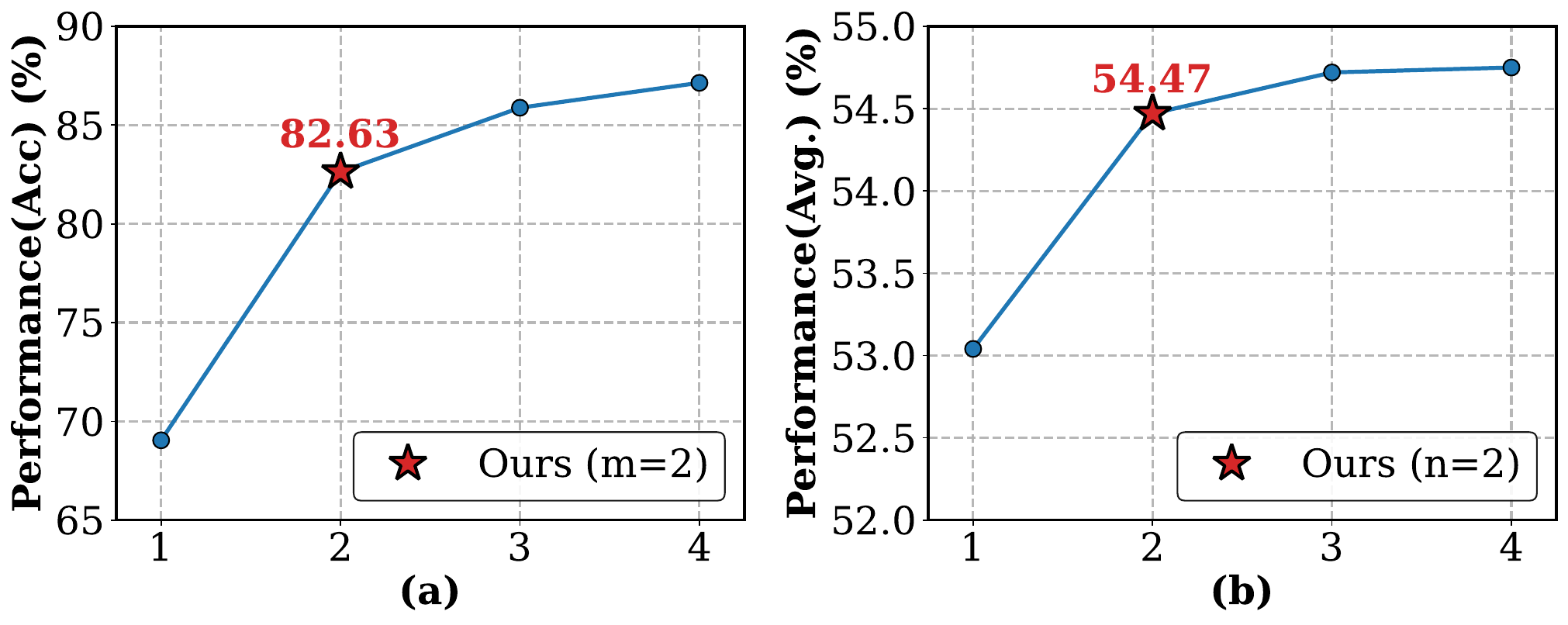}
\caption{Performance under different test-time scaling parameters. (a) Increasing CoT paths $m$ improves diagnosis accuracy, and (b) more segmentation masks $n$ enhances segmentation performance.}
\label{fig:tts}
\end{figure}

As shown in Figure \ref{fig:tts} (a), we investigate the impact of the parameter $m$ denoted in Equation \ref{equ:tts_m} on model performance. The results demonstrate that higher values of $m$ (generating more CoTs) enhance the performance of medical diagnosis.
As illustrated in Figure \ref{fig:tts} (b), we explore the effect of parameter $n$ defined in Equation \ref{equ:tts_n}. The findings reveal that increasing $n$ (producing more segmentation masks) improves medical segmentation performance.

\subsubsection{Generalization Capability for Cross-Modality.}

\begin{table}[t]
\centering
\caption{Cross-modality generalization via testing on untrained modalities. Sim4Seg achieves superior performance, demonstrating robust adaptation to diverse modality of medical data.}
\label{tab:modality_generalization}
\setlength{\tabcolsep}{6pt}
\begin{tabular}{l|cc|cc}
\toprule
\bf \multirow{2}{*}{Modality} &  \multicolumn{2}{c|}{\textbf{LISA}} & \multicolumn{2}{c}{\textbf{Sim4Seg}} \\ \cmidrule(lr){2-3}  \cmidrule(lr){4-5}     & \bf gIoU & \bf cIoU  & \bf gIoU & \bf cIoU   \\
\midrule
X-Ray & 6.34  & 14.01 & \textbf{10.45}  & \textbf{19.07}     \\
Dermoscopy & 43.51  & 27.10 & \textbf{44.97}  & \textbf{34.96}    \\
Endoscopy & 36.13  & \textbf{35.86} & \textbf{41.38}  & 31.12    \\
Ultrasound & 26.17  & 23.42 & \textbf{27.08}  & \textbf{23.69}    \\
Fundus Photography & 13.37  & 13.56 & \textbf{15.76}  & \textbf{16.17}    \\
\bottomrule
\end{tabular}
\end{table}
To validate the modality generalization capability of Sim4Seg, we cyclically excluded the training data of each modality one at a time, trained the model on remaining modalities, and tested on the excluded modality.
As shown in Table \ref{tab:modality_generalization}, our Sim4Seg model demonstrates superior modality generalization performance compared to others. This indicates the robustness and flexibility of our model in adapting to diverse medical imaging modalities.

\subsubsection{Generalization Capability for Untrained Dataset.}

\begin{table}[t]
\centering
\caption{Cross-dataset generalization performance of Sim4Seg when evaluated on datasets excluded from training.}
\label{tab:dataset_generalization}
\setlength{\tabcolsep}{9pt}
\begin{tabular}{l|cc|cc}
\toprule
\bf \multirow{2}{*}{dataset} &  \multicolumn{2}{c|}{\textbf{LISA}} & \multicolumn{2}{c}{\textbf{Sim4Seg}} \\ \cmidrule(lr){2-3}  \cmidrule(lr){4-5}     & \bf gIoU & \bf cIoU  & \bf gIoU & \bf cIoU   \\
\midrule
ISIC & 58.54  & 54.12 & \textbf{60.44}  & \textbf{56.83}    \\
Kvasir-SEG & 36.13  & \textbf{35.86} & \textbf{41.38}  & 31.12    \\
TN3K & 29.06  & 25.40 & \textbf{33.32}  & \textbf{33.62}    \\
ChestX-Det & 6.21  & 7.24 & \textbf{7.78}  & \textbf{9.13}    \\
FIVES & 13.37  & 13.56 & \textbf{15.76}  & \textbf{16.17}    \\
\bottomrule
\end{tabular}
\end{table}
We iteratively excluded the ISIC, Kvasir-SEG, TN3K, ChestX-Det, and FIVES datasets from training data and tested our Sim4Seg model on excluded datasets to evaluate its generalization performance across various medical datasets.
As presented in Table \ref{tab:dataset_generalization}, Sim4Seg achieves better performance in cross-dataset generalization compared to other method, which enhances its real-world applicability in clinical deployment scenarios.
\subsection{Case Study}

Figure \ref{fig:case} presents a multi-modal input case from the M3DS dataset, comparing outputs derived from the baseline model and our approach. 
Compared with the baseline model, our method generated precise segmentation masks along with diagnosis reasoning CoT. This validates the effectiveness of our proposed Sim4Seg model in MDS tasks.
\section{Conclusion}

In this paper, we introduced a new task named Medical Diagnosis Segmentation (MDS). To enable research in MDS, we constructed the Multimodal Multi-disease Medical Diagnosis Segmentation (\textbf{M3DS}) dataset. We further proposed \textbf{Sim4Seg}, an effective model leveraging a novel Region-Aware Vision-Language Similarity to Mask (\textbf{RVLS2M}) module. Additionally, we explored a test-time scaling strategy for MDS task to improve overall performance.

\section{Acknowledgments}
This work was supported by the National Natural Science Foundation of China (No.
624B2002) and the Jiangyin Hi-tech Industrial Development Zone under the Taihu Innovation Scheme (EF2025-00003 SKL-IOTSC).

\clearpage

\appendix
\section{Related Work}

\subsection{Medical Image Segmentation}
Traditional medical image segmentation has relied on fully supervised task-specific architectures~\cite{DBLP:conf/miccai/DrozdzalVCKP16,isensee2018nnu,DBLP:conf/miccai/RonnebergerFB15},
which significantly limits their real-world applications. Recently, to address the growing demand for interactive and context-aware segmentation models in medicine, frameworks supporting various prompts have emerged in the medical
domain ~\cite{MedSAM,DBLP:journals/corr/abs-2308-16184,DBLP:journals/tmi/RajchlLOKPBDRHK17,DBLP:journals/artmed/ZhangSSWZGS21}. However, these models still lack semantic language understanding and corresponding disease diagnosis capabilities. BiomedParse ~\cite{DBLP:journals/corr/abs-2405-12971} directly uses language prompts to predict object shapes and locations, but cannot combine segmentation with diagnostic reasoning, and its prompt support is restricted to class names. While some recent works~\cite{DBLP:journals/corr/abs-2504-11008,DBLP:journals/corr/abs-2505-11872,DBLP:journals/corr/abs-2404-00578} have begun to explore language-guided medical image segmentation and spatial relationship reasoning, their ability to deliver diagnosis results with precise segmentation has yet to be explored.

\subsection{Language-Guided Semantic Segmentation}
Recent advances in Large Multimodal Models (LMMs)~\cite{DBLP:conf/nips/0001WZCY24,DBLP:conf/nips/WangCCWZZLLZQD23,DBLP:conf/nips/WuZXL00WZLL0QD24,DBLP:conf/cvpr/LuCL0KMHK24,DBLP:journals/corr/abs-2502-09838,DBLP:conf/aaai/HuangSLSLHY25} have extended LVLM capabilities to various downstream tasks. For segmentation, one approach trains models to generate polygon coordinates ~\cite{DBLP:conf/nips/WuZXL00WZLL0QD24} or text sequences ~\cite{DBLP:conf/iclr/LanC0XKWF025} as segmentation masks, treating a segmentation model as a mask refiner. While this method directly utilizes LVLMs without additional modules, it often produces verbose text outputs. Alternatively, another method~\cite{DBLP:conf/cvpr/LaiTCLY0J24} uses features from hidden layers of LVLMs to improve the capacity of segmentation models.
Recently, subsequent studies on reasoning segmentation have emerged, including techniques that use points derived from hidden layers as prompts~\cite{qian2024reasoning}, applications to multi-turn dialogue~\cite{DBLP:journals/corr/abs-2312-17240}, and multiple-target segmentation~\cite{DBLP:conf/cvpr/XiaHHPSH24}.
Taking advantage of the two approaches, our proposed framework Sim4Seg leverages features from LVLM hidden layers to generate RVLSMs and employs a segmentation model as a trainable mask refiner.
\section{Detailed Experimental Results of Figure \ref{fig:top_n}.}
Table \ref{tab:TopN} and Table \ref{tab:NtimesN} reveals the detailed experimental results of Figure 6. Table \ref{tab:TopN} shows that performance improves when selecting 12 to 36 grid cells with highest similarity, but declines at 48 grid cells. Similarly, Table \ref{tab:NtimesN} indicates that optimal results emerge when selecting the top 36 grid cells at $16\times 16$ grid resolution, with performance increasing from $8\times 8$ to $16\times 16$ grid resolution and decreasing from $16\times 16$ to $64\times 64$ grid resolution.
In conclusion, excessively coarse grids produce blurred regions, while overly fine grids retain spurious correlations from high-similarity points. These findings demonstrate the critical role of optimal $\tau$ strategy selection in RVLSM effectiveness.
\begin{table}[ht]
\centering
\caption{Impact of selecting 12 to 48 grid cells with highest similarity. Selecting 36 grid cells achieves best performance.}
\label{tab:TopN}
\setlength{\tabcolsep}{14pt}
\begin{tabular}{lccc}
\toprule
\bf top-n & \bf gIoU & \bf cIoU & \bf Avg. \\
\midrule
n=12 & 36.43  &41.78  &39.11\\
n=24 & 46.28  &45.42  &45.85\\
 \rowcolor{gray!15}\bf n=36 & \bf 51.86  &\bf 53.90  &\bf 52.88\\
n=48 & 36.54  &49.73  &43.14\\
\bottomrule
\end{tabular}
\end{table}

\begin{table}[ht]
\centering
\caption{Performance various from $8\times 8$ to $64\times 64$ grid resolution. $16\times16$ grid resolution achieves best performance.}
\label{tab:NtimesN}
\setlength{\tabcolsep}{12pt}
\begin{tabular}{lccc}
\toprule
$\boldsymbol{n} \boldsymbol{\times} \boldsymbol{n}$ & \bf gIoU & \bf cIoU & \bf Avg. \\
\midrule
$8 \times 8$ & 28.61	&37.46	&33.04\\
 \rowcolor{gray!15}$16 \times 16$ & \bf  51.86	&  53.90 & \bf 52.88\\
$32 \times 32$ & 38.98	&\bf 61.55	&50.27\\
$64 \times 64$ & 36.95	&58.14	&47.55\\
\bottomrule
\end{tabular}
\end{table}

\section{Prompt}

\begin{tcolorbox}[
colback=myblue!5!white,
colframe=myblue!75!black,
arc=1mm, 
auto outer arc,
title={Prompt for Medical Assistant},
breakable
]\small

Question: $\{question\}$

Query: You're a medical assistant. Please generate Chain-of-Thought answer in the following format by thinking step by step...

Firstly, the modality of the image is $\{modality\}$

Secondly, I will start to analyze the image ...

Then, ...

...

Finally, $\{answer\}$

\end{tcolorbox}

\begin{tcolorbox}[
colback=myblue!5!white,
colframe=myblue!75!black,
arc=1mm, 
auto outer arc,
title={Prompt for Critical Assistant},
breakable
]\small

Input: $\{CoT\}$

As a rigorous critical assistant, evaluate whether the diagnostic Chain-of-Thought generated by the Medical Assistant is valid.  Assess strictly against these criteria:

Review Dimensions:

1. **Step Completeness**

- Explicitly identifies imaging modality.

- Progressively analyzes key image features.

- Derives final diagnosis through medical reasoning.

2. **Logical Rigor**

- No contradictions/jumps in reasoning.

- Image features substantiate the diagnosis.

- Rules out differential diagnoses (when applicable).

3. **Medical Reliability**

- Terminology conforms to medical standards.

- Diagnosis aligns with current medical consensus.

- No unverifiable/fabricated medical claims.

Output Specifications:

1. **APPROVED** (if all pass):

- Please output [pass]

2. **REJECTED** (if any failure):

- Please output [reject]

Classify failure type:

`Missing Step` | `Logical Flaw` | `Factual Error` | `Non-standard Terminology`

Final decision: [pass]/[reject]

\end{tcolorbox}

\section{Detailed Case Study}

Figure \ref{fig:model_archi} presents multi-modal inputs of each modality from the M3DS dataset, outputs derived from our proposed Sim4Seg model, and the corresponding ground truths. 
Detailed output results of Sim4Seg indicate that our method generated precise segmentation masks with its corresponding diagnosis reasoning CoT. This validates the effectiveness of our proposed Sim4Seg model in MDS tasks.

\begin{figure*}[t]
\centering
\includegraphics[width=0.85\linewidth]{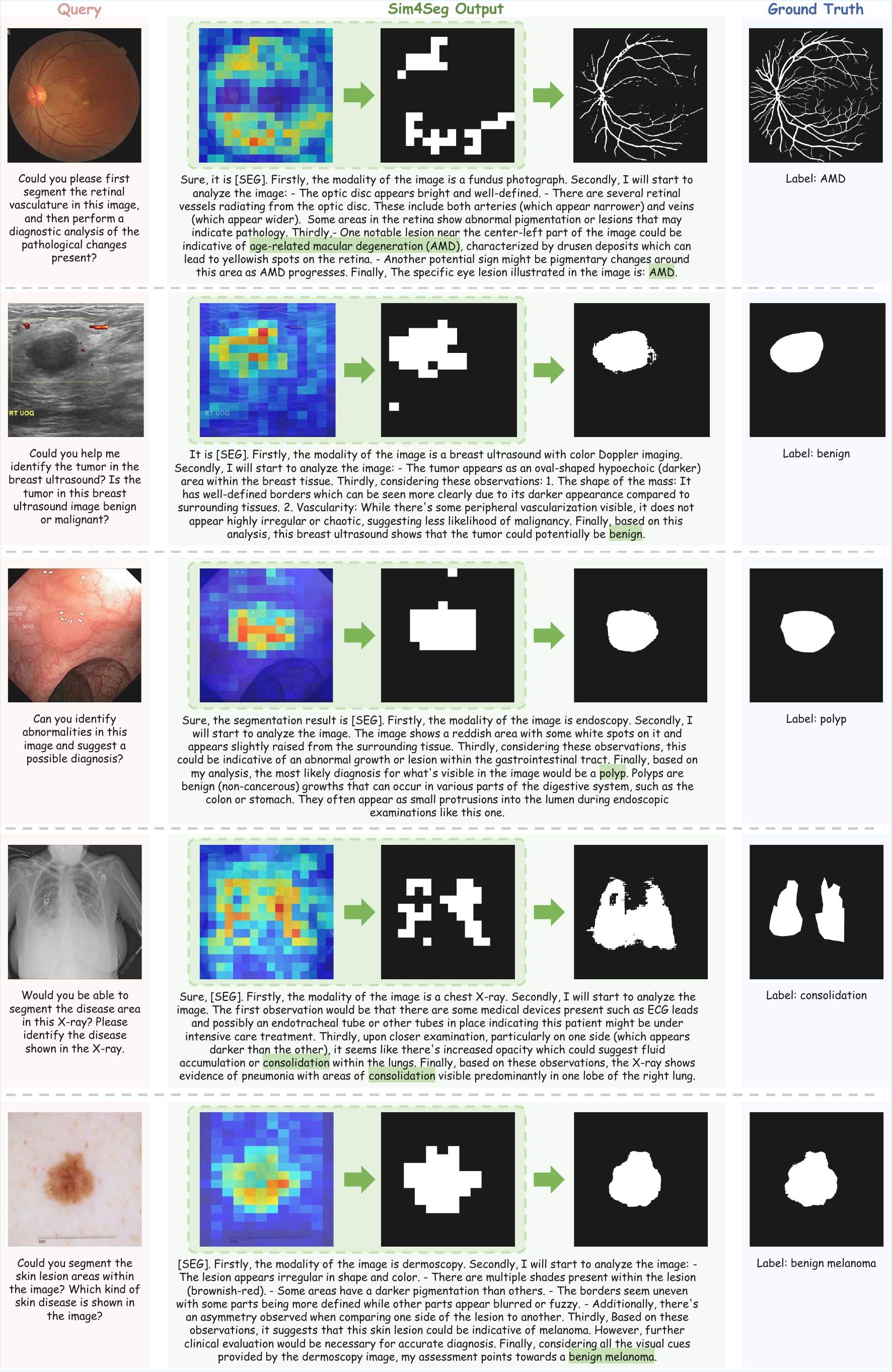}
\caption{\small Detailed Case Study of Sim4Seg. The modalities of the images from top to bottom are: Fundus Photography (FP), Ultrasound (US), Endoscopy (End), X-Ray, and Dermoscopy (DS).}
\label{fig:model_archi}
\vspace{-5mm}
\end{figure*}

\clearpage
\bibliography{aaai2026}

\end{document}